\documentclass{article}

\usepackage[preprint]{neurips_2020}
\usepackage[utf8]{inputenc} 
\usepackage[T1]{fontenc}    
\usepackage{hyperref}       
\usepackage{url}            
\usepackage{booktabs}       
\usepackage{amsfonts}       
\usepackage{nicefrac}       
\usepackage{microtype}      
\usepackage{graphicx}
\usepackage{hyperref}
\usepackage{algorithm}
\usepackage{amsmath}
\usepackage{algorithmic}
\usepackage{makecell}
\usepackage[algo2e]{algorithm2e}
\title{Revisiting Neural Architecture Search}

\author{%
  Anubhav Garg\thanks{Correspondence to \texttt{anubhgar@cisco.com}}  \\
  Cisco Systems\\

  \AND
   Amit Kumar Saha \\
   Cisco Systems\\

  \And
   Debo Dutta \\
   Cisco Systems\\

}

\begin{document}

\maketitle
\begin{abstract}
  Neural Architecture Search (NAS) is a collection of methods to craft the way
  neural networks are built. Current NAS methods are far from ab initio and
  automatic, as they use manual backbone architectures or micro building blocks
  (cells), which have had minor breakthroughs in performance compared to random baselines.
  They also involve a significant manual expert effort in various components of
  the NAS pipeline. This raises a natural question - \emph{Are the current NAS
  methods still heavily dependent on
  manual effort in the search space design and wiring like it was done when
  building models before the advent of NAS?} 
  In this paper, instead of merely chasing slight improvements over 
  state-of-the-art (SOTA) performance, 
  we revisit the fundamental approach to NAS and propose a novel approach
  called $\mathrm{ReNAS}$ that can search for the complete neural
  network without much human effort and is a step closer towards AutoML-nirvana.
  Our method starts from a complete graph mapped to a neural network and searches
  for the connections and operations by balancing the exploration and exploitation of the search space. The results are on-par with the SOTA
  performance with methods that leverage handcrafted blocks. 
  We believe that this approach may lead to newer NAS strategies for a variety of network types. 
\end{abstract}


\section{Introduction}
\label{intro}
Deep Learning has proven to be successful in many application areas such as image 
classification, object detection, semantic segmentation, NLP etc. Much of the current success
is through large datasets and powerful, often handcrafted, neural network models. It is hard
to construct or search for new models ab initio.  Neural Architecture Search (NAS) refers to
finding the neural network for a given task automatically, instead of handcrafting the 
building blocks or layers of the model. It has a great potential to simplify the 
trial-and-error approach of manually designing neural networks, 
can be adapted to work with different compute backends and for a varied
number of tasks. NAS methods have used Reinforcement Learning (\citet{nasnet, metaqnn}), 
Evolutionary algorithms (\citet{amoebanet}), Graph Hypernetworks (\citet{ghypernetworks}) 
while requiring enormous computation resources. More recent NAS methods like ENAS 
(\citet{enas}) and differentiated architecture search methods (\citet{darts, proxylessnas}) 
have reduced the search time to a few GPU days on many different tasks like image 
classification and language modelling. However, current NAS methods still do not achieve full
automation and have the following limitations -

\begin{enumerate}
\item The search space is based on human-made state-of-the-art
networks and thus is already optimal. Hence, a prior is introduced in the architecture 
resulting in similar performance as the random search baselines. The standard deviation in 
accuracy of architectures in such a search space is very low.
\item In most existing NAS methods, the search is done on a micro search space to find the 
most optimal building block or \textit{cell}, using a much smaller network. The best cell is 
then stacked according to the size of the dataset to form a larger network. The improvements 
this approach bring are due to the cell engineering rather than the search method itself. 
That is, the architecture of the cell itself is designed such that any network formed will be
good. This also leads to another manual step and additional hyperparameters, thus making such
methods far from a generic full-fledged architecture search.
\end{enumerate}

The above mentioned shortcomings do not take full advantage of the resources and find the 
structures or patterns which are not known in the literature.
Also, too much effort is spent in improving the SOTA performance instead of 
decreasing the human insight of the network (e.g. starting cell structures etc). 
In this paper, we take a different view and propose a new approach towards 
Neural Architecture Search, a differentiated architecture search method 
that {\em searches for the optimal connections as well as the operations of the neural 
network by varying the exploration vs exploitation trade-off.}
Our goal is slightly different than merely pure SOTA performance.  

Several recent works have shown the inefficiency of the DARTS (\citet{darts}) based search 
space. \citet{hard} showed that the hand-designed cell structure is more 
important than the micro structure (operations) and these cell architectures have a very 
narrow accuracy range. By sampling many random architectures (including the connections 
sampling as well), the accuracy of all the architectures were similar and doesn't degrade 
much when the search space is reduced with inefficient operations. Similarly, 
\citet{randwire} showed that optimal wiring can beat many NAS based architectures without 
searching for the operations. \citet{Yu2020Evaluating} observed that the search policies of different NAS algorithms perform similarly if not worse with random search due to a constrained search space. \citet{understanding-nas} also discover that with the same 
connection topologies as the popular NAS cells but different operations, all random variants 
achieve nearly the same convergence as the popular NAS architectures. This highlights the 
importance of connections in the architecture search as opposed to the current NAS methods,
which are inhibited by the fact that the underlying cell has a strong prior 
(e.g., every node has exactly two input nodes). This makes the network space 
restrictive and explains why random policy has been as effective as sophisticated NAS algorithms.
Hence, it is not clear whether the advantage one NAS method show over another is due to the 
search itself or because of {\em lucky} initialization (\citet{lottery}) or the training protocol. In 
this paper, we argue that a more principled approach for NAS is required to move the space. 

Our method, named $\mathrm{ReNAS}$,
searches for a complete neural network with no predefined backbone, wiring pattern
and operations. We start with a complete graph, mapping it to a neural network: a node
representing a transformation like convolution or pooling and edges representing information
flow. Every node's channels are clustered in blocks. Instead of forming layer-wise connection
pattern, we connect block of channels for each node in our network to every other node block.
This makes our method search on a much larger space than layer-wise connections, previously
unexplored by human experts. It also increases the search speed unlike
DNW~(\citet{wortsman2019discovering}) where every channel is connected to another one making the
search computationally very expensive. Our proposed ReNAS method is based on the weight sharing
differentiable architecture search. 

Our work can be seen as an application of differentiable 
search. Previous work in discovering neural wiring (\citet{wortsman2019discovering})
worked on a MobileNet backbone based on a static structure as the dynamic case is too expensive to 
optimize. \citet{randwire} explore different wiring patterns through a random graph generator
but their wiring is also fixed and is not prior free. To the best of our knowledge, our work is the first step towards full 
automation of the architecture construction without a predefined backbone or cell. 
It removes the manual expert effort in designing various components of the NAS pipeline. 

\section{Related Work}
\label{related}
Here we discuss prior work and their shortcomings and specify how our work overcomes them.

\paragraph{Neural network wiring and topology:} Different types of wiring neural networks have been studied extensively. Inception nets
(\citet{szegedy2015going}) uses modules having parallel paths with different
filters and concatenates them. ResNet (\citet{he2016deep}) uses shortcut or skip 
connections to learn the residual function $\mathcal{F}$($x$) $+$ $x$, while 
DenseNet (\citet{Huang2017DenselyCC}) uses connections between every pair of 
layers in a block. MaskConnect (\citet{maskconnect}) learns the connection 
between modules in a residual network by assigning each connection a real valued 
weight. They chose K (hyperparameter) input connections for each module and learn
the connection weights along with network parameters. \citet{randwire} uses 
stochastic network generator to generate a graph which is mapped to a neural 
network (RandWire). The wiring in RandWire is fixed by the network generator's 
prior (WS, ER, BA) and the seed. \citet{Yuan2019DivingIO} uses complete graphs in
stages and learns the connection weights in a stage by continuous relaxation as 
in DARTS. However, the learned architecture is never discretized at the end of 
training, resulting in a DenseNet style architecture with weighted incoming 
connections instead of a concatenation of all connections. Our work does not search
for layer-wise connections, instead finds connections between channel partitions.
\citet{wortsman2019discovering} relaxes the notion of layers and learn the wiring
between channels. However, due to their expensive optimization, it is constrained 
to generating smaller networks in low computation regime. Our work is motivated 
by finer connections than layer or node, but instead of allowing all channel wise
connections, which is more computationally and memory expensive, uses blocks of 
channel connections.

\paragraph{Neural Architecture Search:} \citet{nasnet} used a controller RNN and trained it
with reinforcement learning to search for architectures. Since then, numerous NAS methods
have been studied. Based on major search strategy, NAS methods can be classified into
Reinforcement Learning (\citet{metaqnn, network-transformation, block-nas, nasnet-cell,
enas}), Neuro-Evolution (\citet{real2017large, suganuma2018exploiting,
hierarchical-evolution, real2019regularized, lamarckian}) and gradient-based (\citet{darts,
proxylessnas, snas, xnas, pcdarts, progressive-darts}). Other methods include Random Search
(\citet{random-search}), Bayesian Optimization (\citet{jin2019auto, bayesian-nas} and some
custom methods (\citet{renas, envelopenet, manas, progressive-nas}). In this work, we search
for the connection as well as the operations. Unlike previous NAS methods, which uses a
backbone or pre-designed cell, and then search on them, we visualize the neural network as a
graph having groups of channel-wise connections. Our method prunes the redundant connections
and search for the operations at the same time.

\section{Method}
\label{method}
In this section, we describe our method for learning the structure as well as the operations 
of a neural network at the same time. We first describe the construction of the 
over-parameterized neural network, then the search space, and finally the method.

\subsection{Constructing the Parent Network}
\label{con-over}
\begin{figure}[t]

\begin{minipage}[t]{.45\textwidth}
  \centering
  \includegraphics[width=50mm,scale=0.5]{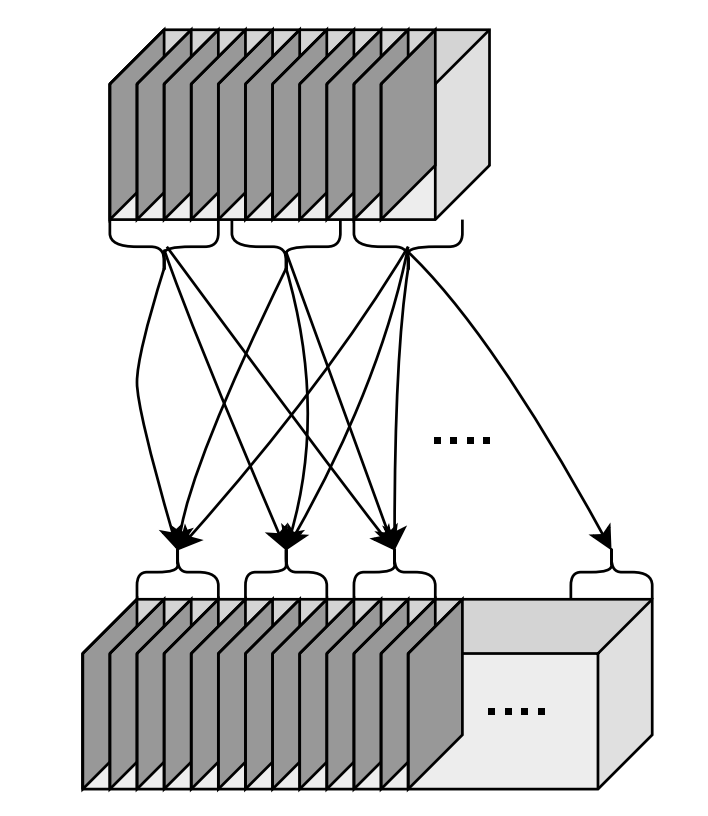}
  \caption{Node level connections. $K$ blocks of one node are connected to $L$ blocks of another node via $K\times L$ connections.}
  \label{fig:connections}
\end{minipage}
\hspace{1.1cm}
\begin{minipage}[t]{.45\textwidth}
  \centering
  \includegraphics[width=50mm,scale=0.5]{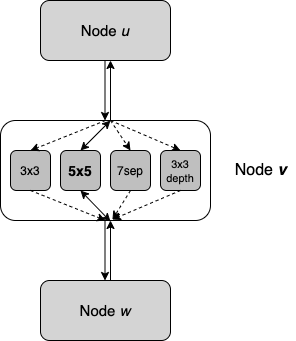}
  \caption{Operation selection step. An operation is sampled from the search space, here a $5\times5$, and it's $\alpha$ is updated acc. to Eq. \ref{eq3} }
  \label{fig:operation}
\end{minipage}

\end{figure}

The searchable parent network consists of multiple directed acyclic graphs (DAG) \( 
\mathcal{G} =  ( \mathcal{V}, \mathcal{E} ) \) where every node $v$ $\in$ \( \mathcal{V} \) 
is assigned an index. There is an edge $e_{ij}$ $\in$ \( \mathcal{E} \) for every node $v_i$ 
to $v_j$, $i<j$. Nodes represent some transformation of the input data (e.g., pooling, 
convolution) and edges represent data flow. A node $v_i$ having $C$ channels can be 
partitioned into $K$ channel blocks. Every channel block Hence, for pair of nodes $v_i$ and node $v_j$, there are
$K^2$ connections. Each edge $e_{ij}$ has weight $\gamma_{ij}$. Each DAG \( \mathcal{G}_n 
\)'s input is the output of the previous DAG \( \mathcal{G}_m \), $m<n$. The input 
featuremap's size of every DAG is reduced by half and the channels are increased to twice by 
the first node of every DAG. Figure~\ref{fig:connections} 
depicts the node level connections in the one-shot network.

\subsection{Search Space}
We search for the complete architecture on target task and dataset, without any proxy, similar to
(\citet{proxylessnas, Hu2020DSNASDN}). Our approach doesn't require any predefined backbone architecture
though, as with previous approaches. For a complete DAG with $N$ nodes, there are $2^{N(N-1)/2}$ possible
connections. For every node in a DAG, there is an operation (e.g. convolution, pooling) associated with it,
thus for $O$ operations, there are total $O^N \times 2^{N(N-1)/2}$ networks, since every choice is independent.
For $M$ such DAG's, there are $M \times O^N \times 2^{N(N-1)/2}$ networks in our search space. This is
significantly larger than earlier methods, which use either building blocks in a backbone architecture or
expert engineered cells which have a restricted search space.

\subsection{Architecture Search}
The architecture search is based upon the differentiable architecture search (\citet{darts, 
proxylessnas}). In this method, we start with an over-parameterized (parent) network 
having all operations in the search space (e.g., convolution, pooling, etc.). Every operation
and edge is assigned a weight parameter which is trainable like the weights of the neural network. Every 
iteration is composed of two steps - 1) The weights $w$ of the neural network are trained by
gradient descent algorithms on a training set keeping the architecture parameters fixed and 2) The architecture
parameters $\alpha$ are trained on a separate validation set according to gradient based methods (SGD or exponential) keeping the 
weights fixed. The parameters $\gamma$ of the connections are trained along with the weight parameters of the parent network. Algorithm \ref{alg:algo} details the steps of our method.

\begin{algorithm}[H]
 \caption{$\mathrm{ReNAS}$}
 \label{alg:algo}
\SetAlgoLined
 Construct the over-parameterized network $\mathbb{G}(w, \alpha, \gamma)$ by initializing every DAG in the network\newline
 \While{not converged}{
  \For{each $\mathcal{G}$ in $\mathbb{G}(w, \alpha, \gamma)$}{
  \For{each node $v$ in \( \mathcal{G}\)}{
  1. Partition the node into $K$ channel blocks\newline
  2. Calculate $x^k$ for $k = 1 ,..., K$ according to Eq. \ref{eq1}\newline
  3. Sample an operation from the Operation Search Space\newline
  4. Calculate the resulting featuremap $X$ for node by concatenating $X^k$ $\forall$ $k$ according to Eq. \ref{eq2} \\
  5. Update the weights $w$ and $\gamma$ using SGD\\
  6. Fix $\gamma$, $w$ and update $\alpha$ based on Eq. \ref{eq3}
  }
  }
 }
 For each node, retain the operations with highest $\alpha$ and half channel block connections with highest $\gamma$. Prune all other connections
\end{algorithm}
\begin{equation}
  \label{eq1}
x^k\, =\, \sum^{K}_{l=1} \, \gamma_{ijl} *X^{k}_i  \hspace{2em} for \, \, k = 1, ... ,K \hspace{2em} \forall \hspace{0.75em} j>i
\end{equation}
\begin{equation}
  \label{eq2}
X^k\, =\,  o_j(x^k) \hspace{2em} for \, \, k = 1, ... ,K
\end{equation}

Here $x^k$ is the intermediate variable, $X_i$ is the featuremap of node $i$ and 
$o_j$ is the sampled operation for node $j$ from the operation search space . 

The procedure starts by constructing the parent network by wiring
$M$ DAG's $\mathcal{G}$ as described in section
\ref{con-over}. Every node's channels in $\mathcal{G}$ are partitioned into $K$ channel blocks.
Here $K$ is a hyperparameter which controls the granularity of 
our search; $K=1$ means there is no partition. If $K=$ number of channel in node
$u_i$ it means that every channel is connected to every other channel and the
wiring of the over-parameterized network is similar to
\citet{wortsman2019discovering}.  Each partition of node $v_i$ can be visualized 
as a separate entity or node. Each channel block of node $v_i$ is connected to every other channel block of node $v_j$, thus forming $K^2$ connections. 
After that, each such connection is weighted through the connection parameter 
$\gamma_{ij}$ according to Eq.~(\ref{eq1}). Then, one of the operations is sampled from the operation search
space and its parameter $\alpha$ is updated. Figure~\ref{fig:operation} depicts the node level expansion. The
operation sampling is based on the path sampling heuristic of (\citet{proxylessnas}), in which two candidate
paths (operations) are sampled from a multinomial distribution over all operations. We update these 
parameters $\alpha$ of the over-parameterized network using the loss gradient equation as described in
\cite{proxylessnas}. 
Here, $\delta_{mn}$ is equal to 
$0$ when $m$ is not equal to $n$ and $1$ when $m=n$. $p_m$ and $p_n$ are the 
probabilities of choosing the operation $m$ and $n$ respectively using the 
multinomial probability distribution on parameters $\alpha$ of operations.

\begin{equation}
   \label{eq3}
\frac{\partial L}{\partial \alpha_{m}} = \sum^{2}_{n=1} \frac{ \partial L}{\partial g_n}p_n(\delta_{mn} - p_m)
\end{equation}

The selected operation $o$ is then applied to the input $x$ to obtain the featuremap of 
node $v_j$. This process is repeated for every node of every DAG in the over-parameterized 
network until convergence. We prune half of the connection parameters $\gamma$ after training
the over-parameterized network. For every node, the operation with highest alpha are 
retained.

Our method allows us to search for more diverse architectures than previous NAS and 
hand-crafted architectures. It is more fine-grained than those methods yet efficient than searching via 
all channel level connections. Also, the exploration and exploitation of the search space can
be controlled by changing the value of $K$, which allows our method to be useful in various 
environments ranging from  $1$ to hundreds of GPUs.

\section{Experiments}
\label{experiments}
We take the target task as image classification in our experiments, 
hence we search for a convolutional neural network directly without cells 
or building blocks. Our experiments are run on NVIDIA Tesla V100 GPUs.
Our operations search space 
consists of $3\times3$, $5\times5$ and $7\times7$ depthwise seperable convolutions and 
$3\times3$, $5\times5$ and $7\times7$ convolutions.

We demonstrate the efficacy of our method on the CIFAR-10 (\citet{cifar10}) image classification benchmark 
dataset. CIFAR-10 dataset consists of 60000 coloured images of size 32X32 split in 50000 training and 10000
test images equally in 10 classes. We sample 5000 images randomly from the 
training data for the training of the operation parameters $\alpha$. 
The weights $w$ and $\gamma$ are updated via SGD optimizer with a momentum of 0.9 
and a cosine learning rate scheduler with initial learning rate of $0.1$. 
For updating $\alpha$, we use the Adam optimizer with an initial learning rate of $0.006$. 
Table~\ref{tab:result} compares our results with 
previous NAS methods. The value of the hyperparameter $K$ is equal to 4 in the experiments. 

\begin{table}[htbp]
    \centering
    \caption{Comparison of different NAS methods}
    \begin{tabular}{l|l|l|l}
    \specialrule{.1em}{.05em}{.05em}
    Method & Params (M) & Test error (\%) & Search time (GPU Days)\\
    \specialrule{.1em}{.1em}{.1em} 
    NASNet-v3 (\cite{nasnet})  & 37.4 & 3.65  & 1800\\
    \hline
    Block-QNN (\cite{metaqnn})  & 39.8 & 3.54 & 96\\
    \hline
    AmoebaNet-B (\cite{amoebanet})  &  34.9 & 2.13  & 3150\\
    \hline
    PNAS (\cite{progressive-nas})  & 3.2 & 3.41  & 225\\
    \hline
    ENAS (\cite{enas})  &  4.6 & 3.54  & 0.45\\
    \hline
    DARTS (\cite{darts}) &  4.6 & 2.76  & 4\\
    \hline
    ProxylessNAS (\cite{proxylessnas})  & 5.7 & 2.08  & 8.3\\
    \hline
    SNAS (\cite{snas})  & 2.85 & 2.8  & 1.5\\
    \hline
    PCDARTS (\citet{pcdarts}) & 3.6 & 2.57  & 0.1\\
    \hline
    \makecell[l]{DNW-MobileNetV1\\
    (\citet{wortsman2019discovering})}& 12.11  & 9.8 & 0.2 \\
    \hline
    ReNAS  & 4.2 & 2.72  & 0.5\\
    \specialrule{.1em}{.05em}{.05em}
    \end{tabular}
    \label{tab:result}
\end{table}

The search time of our method $\mathrm{ReNAS}$ with other details is specified
in the last row. We note that like other methods, $\mathrm{ReNAS}$ is a 
two-stage approach, having a search and retrain stage. The table shows that
ENAS~(\citet{enas})
has a slightly lower search time but has higher error rate (lower accuracy)
in spite of having more parameters.
Similarly, ProxylessNAS~(\citet{proxylessnas}) has a slightly better accuracy
but uses considerably higher search time to come up with higher number of parameters.

We note that while PCDARTS~(\citet{pcdarts}) has the least search time with slightly greater 
accuracy, it is a more complex technique well designed for the already optimal
DARTS search space. As discussed in Section~\ref{intro}, this is a restricted 
search space with a very narrow accuracy range. For example, \citet{hard} 
found out that from $214$ sampled architecture from this search space, all 
perform similarly with a mean accuracy of $97.03 \pm 0.23$. 
The main purpose of this paper is not to introduce another expert
designed method for a niche space and the
goal is not to beat the state-of-the-art, but to design a principled approach 
to finding interesting structures and patterns through automation.

We also compare 
$\mathrm{ReNAS}$ with DNW~(\citet{wortsman2019discovering}), as it unifies 
core parts of the sparse neural network literature with the neural 
architecture search problem. Strictly speaking, DNW is not a NAS method
and takes a backbone architecture like MobileNet 
and discovers the edge connections, but it has a constraint that the total number of 
edges at every round is limited by a hyperparameter $k$. 
In our method, we do not have such a restriction on 
the number of edges and explore all block level connections without any backbone architecture and
simultaneously search for the operations. 
Even though the proposed method has a significantly large 
search space, its performance is on-par with other NAS methods.

\section{Discussion}
\label{discussion}
Our method is able to control the granularity of the architecture search. We have observed 
that the current deep learning frameworks are able to optimize the computation in a layer 
wise neural network. Our method is not based on layer-wise architecture and instead has node 
level connections. Hence, the weight computations between nodes in such an architecture does 
not take advantage of the many tensor optimizations in current frameworks. This is one of the
reason why our method is not able to beat the current SOTA methods though it is on par with 
those methods. Our method should be able to learn complex connections and patterns which are 
not prevalent in the current literature.
We expect it to perform better on unseen and unconventional datasets, thus making this approach
applicable to a wider set of problems.

\section{Conclusions}
\label{conclusions}
In this paper, we take a fresh look at NAS, beyond chasing SOTA performance. 
We recognize various problems in the current NAS methods, e.g., the problem of searching within 
a limited space with hand crafted cells and manual backbones and present a novel solution 
using a gradient-based approach. We use a finer-grained
search by having channel-wise block connections, which is better than 
searching all channel-to-channel connections. Searching channel level connections 
instead of the block-of-channel connections is expected to provide a more fine grained result, 
which is a part of our future work. The preliminary results, shown in this paper, 
are promisingly close to SOTA, and we believe that this pattern will apply to other NAS problems. 

\bibliography{references}
\bibliographystyle{neurips_2020}

\end{document}